\documentclass[letterpaper, 10 pt, journal, twoside]{IEEEtran}

\usepackage{hyperref}
\usepackage{graphicx}

\ifCLASSINFOpdf
\else
\fi
\usepackage{amsmath}
\usepackage{upgreek}
\usepackage[dvipsnames]{xcolor}
\newcommand{\newtext}[1]{\textcolor{black}{#1}}

\usepackage{titlesec}

\titlespacing*{\subsection}{1pt}{1pt}{1pt}
\titlespacing*{\section}{2pt}{2pt}{2pt}

\begin{document}
%
\title{Finite Element Modeling of Pneumatic \\ Bending Actuators for Inflated-Beam Robots}
%
%
%
%

\author{Cosima~du Pasquier,
        Sehui~Jeong,
        and~Allison~M.~Okamura,~\IEEEmembership{Fellow,~IEEE}%
\thanks{Manuscript received: June, 6, 2023; Revised August, 21, 2023; Accepted September, 17, 2023.}
\thanks{This paper was recommended for publication by Editor Editor Yong-Lae Park upon evaluation of the Associate Editor and Reviewers' comments. \\
~\copyright~2023 IEEE.  Personal use of this material is permitted.  Permission from IEEE must be obtained for all other uses, in any current or future media, including reprinting/republishing this material for advertising or promotional purposes, creating new collective works, for resale or redistribution to servers or lists, or reuse of any copyrighted component of this work in other works.\\
IEEE policy provides that authors are free to follow funder public access mandates to post accepted articles in repositories. When posting in a repository, the IEEE embargo period is 24 months. However, IEEE recognizes that posting requirements and embargo periods vary by funder. IEEE authors may comply with requirements to deposit their accepted papers in a repository per funder requirements where the embargo is less than 24 months. Information on specific funder requirements can be found \href{https://open.ieee.org/for-authors/funders/}{here}. \\
The work was supported in part by the U.S. Department of Energy, National Nuclear Security Administration, Office of Defense Nuclear Nonproliferation Research and Development (DNN R\&D) under subcontract DE-AC02-05CH11231 from Lawrence Berkeley National Laboratory, and National Science Foundation grant 2024247.} 

\thanks{All authors are with the Department of Mechanical Engineering, Stanford University, United States
        {\tt\footnotesize cosimad@stanford.edu}}%
\thanks{Digital Object Identifier (DOI): see top of this page.}
}

%
%

\markboth{IEEE ROBOTICS AND AUTOMATION LETTERS. PREPRINT VERSION. ACCEPTED SEPTEMBER 2023}
{du Pasquier \MakeLowercase{\textit{et al.}}: Finite Element Modeling of Pneumatically Bending Actuators for Inflated-Beam Robots}
%

\IEEEpubid{\makebox[\columnwidth]{\hfill 10.1109/LRA.2023.3320010~\copyright~2023 IEEE} \hspace{\columnsep}\makebox[\columnwidth]{}]}


\IEEEtitleabstractindextext{%
\begin{abstract}
\newtext{Inflated-beam soft robots, such as tip-everting vine robots, can control curvature by contracting one beam side via pneumatic actuation. This work develops a general finite element modeling approach to characterize their bending. The model is validated across four pneumatic actuator types (series, compression, embedded, and fabric pneumatic artificial muscles), and can be extended to other designs. These actuators employ two bending mechanisms: geometry-based contraction and material-based contraction. The model accounts for intricate nonlinear effects of buckling and anisotropy. Experimental validation includes three working pressures (10, 20, and 30 kPa) for each actuator type. Geometry-based contraction yields significant deformation (92.1\% accuracy) once the buckling pattern forms, reducing slightly to 80.7\% accuracy at lower pressures due to stress singularities during buckling. Material-based contraction achieves smaller bending angles but remains at least 96.7\% accurate. The open source models available at http://www.vinerobots.org support designing inflated-beam robots like tip-everting vine robots, contributing to waste reduction by optimizing designs based on material properties and stress distribution for effective bending and stress management.}
\end{abstract}

}

\maketitle

\IEEEdisplaynontitleabstractindextext

%
\IEEEpeerreviewmaketitle

\ifCLASSOPTIONcompsoc
\IEEEraisesectionheading{\section{Introduction}\label{sec:introduction}}
\else
\newtext{\section{Introduction}}
\label{sec:introduction}
\fi

\IEEEPARstart{S}{oft} robotic systems allow for safe physical interaction with complex or delicate environments. In contrast to to their rigid counterparts, of soft robots use embedded design features, \newtext{such as actuators or pleats, linking} form to function \cite{Sparrman2021PrintedRobotics}. Typically, refining soft robot designs relies on lengthy iterative experimentation. Yet, finite element (FE) modeling streamlines design optimization and performance prediction, reducing the material and time cost for new soft robots while ensuring safe and reliable operation \cite{duPasquier2019DesignRobotics,Duriez2013ControlMethod}. \par
Inflated beam robots (IBRs) are a class of soft robots that use pressurized textile or plastic sleeves to achieve a range of shapes and access constrained or cluttered environments \cite{Blumenschein2022GeometricRobots,Jitosho2023PassiveRobots,Hwee2021AnDevice}. With multiple degrees of freedom (DoFs), IBRs re controlled through pleats \cite{Voisembert2013DesignActuation}, internal devices \cite{Takahashi2021EversionFunction,Haggerty2021HybridCapabilities}, tendons \cite{Blumenschein2022GeometricRobots,Gan20203DRobots}, and pneumatic actuators \cite{Coad2020VineExploration,Naclerio2020SimpleMuscle,Niiyama2015PouchDesign}. \par
Tip-everting inflated beam robots, or vine robots, are a subset of IBRs in which the beam wall material is initially inverted inside the robot. Pressure-driven eversion causes the vine robot to “grow”, extending its tip while the beam wall is stationary relative to its environment. This is advantageous for applications with delicate environments to be navigated via tortuous paths, like surgery, search and rescue, and archaeology \cite{Coad2020VineExploration,Hawkes2017AGrowth}. Challenges arise in beam bending due to sleeve and eversion limits, prompting the design of pneumatic actuators that remain compact during eversion and then expand to control beam shape \cite{Kubler2023ASteering}. \\
Previous work on modeling of pneumatic actuators for IBRs focused on kinematics and geometric approximations of volume change, virtual work, and conservation of energy \cite{Niiyama2015PouchDesign,Greer2017SeriesRobot,Abrar2021HighlyStructure}. These models only offer a rough approximation of the deformed center axis of the IBR. 
Pneumatic actuators rely on geometry-based contraction or material-based contraction for bending. Geometric contraction causes distributed buckling throughout the beam influenced by material and stress distributions, not captured by kinematic models. Material-based contraction relies on material anisotropy that is shear stress distribution dependent, also not captured by kinematic modeling. Thus, existing models only approximate deformation through iterative parametric tuning based on experimental data, requiring time-intensive prototyping and testing procedures. Each change in design or material requires a new tuning process. \\
In contrast, finite element modeling (FEM) captures internal forces, strains, and stresses, and can predict both buckling and anisotropy. FEM requires initial material characterization and mesh refinement, but adapts to geometry and material changes, seamlessly including external loads. \\   
\newtext{In this work, we contribute to the field of soft robotics in three distinct ways. First, we propose the first comprehensive and open source FE framework for IBRs (available at http://www.vinerobots.org), predicting bending mechanisms incorporating both buckling and anisotropy. The framework can be modified to simulate and verify bending in other pneumatic actuators for IBRs, streamlining their design and fabrication.  Then, we provide constitutive models for quasi-isotropic and inextensible materials used for buckling and anisotropic extensible materials used for anisotropy, integrated into the FE software Abaqus CAE \cite{Smith2014ABAQUS/Standard6.14}. Third, by utilizing \emph{Dynamic, Explicit}\footnote{Abaqus CAE-specific functionalities are capitalized and italicized throughout the text.} FE analysis, our data-driven model predicts deformation for four main types of actuators: series, compression, fabric, and embedded pneumatic artificial muscles (sPAMs \cite{Coad2020VineExploration}, cPAMs \cite{Kubler2023ARobots}, fPAM \cite{Naclerio2020SimpleMuscle}, ePAMs \cite{Abrar2021HighlyStructure}). sPAMs are also referred to in literature as series Pouch Motors, so we refer to them here as sPAMs/PMs \cite{Niiyama2015PouchDesign}.} \\
This paper is organized as follows: first, we describe the material mechanical characterization and the benchmark experimental bending protocol for all actuators. Then, we explain the assumptions for the FE model mirroring the experimental setup. We define comparison metrics and error calculations for FE model validation. Next, we present experimental and simulation bending results and discuss the validity and robustness of our approach. Finally, we discuss the results in the broader context of pneumatic actuation of IBRs. \newline

\newtext{\section{Background}}
\subsection{Modeling of Deformation in Pneumatically Actuated IBRs}
The complexity of controlling an IBR with pneumatic actuators, unlike other mechanisms such as internal devices \cite{Takahashi2021EversionFunction} or tendons \cite{Blumenschein2022GeometricRobots} arises from the fact that the relationship between actuation and shape cannot be directly extrapolated from finite orientation or length changes. Differences in bending mechanisms among pneumatic actuator types also impact their modeling. \par Pneumatic actuators induce bending by shortening one side of an IBR, achievable through geometry-based contraction, newtextwhere the transition from flat to 3D inflated configurations shortens the actuator, or material anisotropy, where the inherent material properties cause the actuators to shorten. \par
The sPAM/PM, cPAM, and ePAM bending actuators all rely on geometry-based contraction, often modeled with energy conservation and virtual work principles \cite{Greer2017SeriesRobot,Abrar2021HighlyStructure,Kubler2023ARobots,Greer2019AExtension}. Analytical models describe geometric changes of actuators during inflation, equating pressure-induced work to virtual translation or contraction. They assume material inextensibility, constant curvature, and idealized linear spring behavior. Niiyama et al. \cite{Niiyama2015PouchDesign} and Greer et al. \cite{Greer2019AExtension} model sPAMs/PMs separately from the IBR body and for low pressures. Kübler et al. \cite{Kubler2023ARobots} model sPAMs/PMs and cPAMs including curvature and show promising results, but the accuracy varies with actuator dimensions. Abrar et al. \cite{Abrar2021HighlyStructure} model bending of ePAMs, but the model and the experimental data differ significantly. \par
fPAMs rely on material anisotropy for bending. Naclerio and Hawkes \cite{Naclerio2020SimpleMuscle} relate pressure increase to volume increase based on McKibben muscle principles. Ultimate tensile strength is used to determine maximum pressure and maximum actuator contraction. The model performs very well for pure linear displacement of isolated actuators. Kübler et al. \cite{Kubler2023ARobots} extend the model to bending by including the reaction forces between the IBR and the actuators in the equilibrium equations. \par
Analytical models for both contraction types are actuator-specific and tuned using experimental data. They lack stress and strain fields, focusing only on shape prediction without considering the full range of information required for actuator design and fabrication. 

\newtext{\subsection{Considerations in FEA of thin-walled structures}}
The body of an IBR is a thin-walled cylindrical shell. The actuators on or in the body surface generate asymmetric lateral loading, which in turn causes the cylinder to bend through buckling. This specific scenario has not been documented in literature for FE analysis, but there is a body of literature concerning loading of thin-walled cylindrical shells under uniform lateral pressure \cite{Tabiei2012NumericalPressure,Rostamijavanani2020DynamicPressures}. \par
There are two methods to solve dynamic buckling FE problems with ABAQUS: \emph{Standard} and \emph{Explicit}. They use two different approaches to solving the general equilibrium equations in a structure:
\begin{equation}
    \mathbf{M} \mathbf{\ddot u} + \mathbf{D} \mathbf{\dot u} + \mathbf{K} \mathbf{u} = \mathbf{F},
\end{equation}
where \textbf{M} is the mass matrix, \textbf{D} is the damping matrix, \textbf{K} is the stiffness matrix, \textbf{F} is the external load vector, and \textbf{u} is the nodal displacement vector. \par
\newtext{In \emph{ABAQUS/Standard}, the Newton-Raphson method solves the equations implicitly for stability. The stiffness matrix (K) is updated for the deformed structure at each step. \emph{ABAQUS/Explicit} uses the central difference formula. While less stable, accuracy can be improved by reducing time step size. Despite \emph{ABAQUS/Explicit} requiring more time steps, in complex problems like ours \emph{ABAQUS/Standard} has a high iteration cost. ABAQUS/Explicit, acknowledged for efficiency in scenarios with frequent self-contact \cite{Smith2014ABAQUS/Standard6.14}, was chosen for our model. Also, \emph{ABAQUS/Explicit} offers the \emph{*Fabric} material model capturing unique traits of woven fabrics. This is beneficial for our study, given our fPAM's anisotropic fabric, challenging to represent with hyperelastic models.} \par

\newtext{\emph{ABAQUS/Explicit} offers two methods to define our time step: \emph{Dynamic} and \emph{Quasi-Static}. The \emph{Quasi-Static} approach omits inertial forces. This method can be effected if calibrated for loading rate and mass scaling to limit inertial forces while keeping simulations relatively quick. Thus, preliminary simulation batches are necessary to estimate these values, adding complexity to FE simulation setup.} \par
\newtext{Previous studies confirm the suitability of the \emph{Explicit, Dynamic} method for simulating thin-walled inflatable structures. Nguyen \& Zhang \cite{Nguyen2020DesignDevices} and Baines et al. \cite{baines2021rapidly} used \emph{Explicit-Dynamic} for fabric soft pneumatic actuators and inextensible material inflatables. Thus, we use \emph{Dynamic, Explicit} to model bending actuators in IBRs. Valid solutions must maintain kinetic energy below 10\% of total internal energy \cite{Smith2014ABAQUS/Standard6.14}.}

\section{Methods}
This section introduces the FE modeling approach and parameters for IBRs, based prior thin-walled structures research. We detail actuator materials, data acquisition methods for constitutive material models, actuator designs, assembly methods, and performance assessment protocol.

 \subsection{FE Model for IBRs}
Using Abaqus CAE \emph{Explicit}, we developed an FE model to predict deformations and reaction forces in IBRs and their actuators. We used a \emph{Dynamic, Explicit} step was. To optimize computation, IBR and actuator models were defined as shells with thicknesses of 200 $\upmu$m and 50 $\upmu$m for TPU-coated and silicone-coated fabrics, respectively. For efficiency, both IBR and actuator meshes were set at 4.0 mm and 1.0 mm respectively, roughly 1\% of the IBR length and 1.7\% of a single actuator length. \newtext{These sizes, determined through a preliminary study spanning mesh sizes from 0.5 mm to 4.0 mm, were chosen for their convergence accuracy and computational efficiency.} \par
The cPAM and ePAM actuators' model employed \emph{Tie} constraints on all welded surfaces. Similarly, sPAMs/PMs utilized Tie constraints for both welded and glued surfaces. To model the fPAM, given distinct elastic properties of the material and glue, the glued surface was represented as a \emph{Composite} surface, combining silicone-coated Nylon material properties with those of the glue. \emph{General Contact} was implemented to prevent collisions, and a 0.01 mm gap separated IBR and actuator layers, ensuring a collision-free model initialization. \par
A maximum pressure of 2 kPa was applied to the IBR, and a maximum pressure of 10, 20, and 30 kPa was applied to all actuator interiors. \newtext{In a single 1.1 second time step, IBR pressure was applied in the first 0.1 seconds, then held (A = 0 at t = 0, A = 1 at t = 0.1, A = 1 at t = 1.1), then actuator pressures were applied for the remaining 1 second (A = 0 at t = 0, A = 0 at t = 0.1, A = 1 at t = 1.1) using two distinct \emph{Smooth Amplitudes}, predefined 5$^\mathrm{th}$-order polynomial curves. The choice of 1.1 seconds balances computational effort while keeping model kinetic energy under 1\% of total energy.} To avoid inertial effect and to ensure that steady-state deformation has been reached, each maximum pressure is simulated separately. \par
Both \emph{Pinned} and \emph{Encastered} boundary conditions (BCs) were tested. The change did not affect the results, so pinned BCs were applied to the final simulations. \par
\subsection{Material Characterization}
The sPAMs/PMs, cPAMs, and ePAMs all rely on geometric shape change for contraction, and thus require a sturdy inextensible material. We chose a 70D TPU-coated Ripstop Nylon because it easily bonds to itself when the coating is heated, which we achieve through ultrasonic welding, and it does not rip at the current working pressures. The fPAM relies on material anisotropy for contraction. We chose a silicone-coated Ripstop Nylon because of its proven anisotropic behavior in previous work \cite{Naclerio2020SimpleMuscle}. The material properties needed for FEA were measured using uniaxial testing to break on an Instron 5565 under ASTM D882 measurement standard and rectangular test specimens \newtext{(20 mm$\times$350 mm, 250 mm gauge length)}. Five specimens were tested per material and per relevant orientation. The measured elastic moduli for the Nylon and the silicone fabrics are shown in Table \ref{Tab:E_moduli} for each orientation. \par
In the experiments, the sPAMs/PMs, cPAMs, and ePAMs and their respective IBRs are all cut along the main axes of the textile (0$^\circ$ and 90$^\circ$). The \emph{Material Evaluation} module on Abaqus CAE was used to fit the 0$^\circ$ and 90$^\circ$ uniaxial test data of the TPU-coated Nylon to the \emph{Reduced Polynomial} hyperelastic constitutive model, \newtext{also known as Neo-Hookean model}, with $N = 1$, where $U = C_{10} (\bar I - 3)$, resulting in the following material parameters: C10 = 50.3, D10 = 0. \par
\begin{table}[h]
\caption{Elastic moduli for both materials used to fabricate four types of IBR actuators. }
\label{Tab:E_moduli}
\renewcommand{\arraystretch}{1.2}
\begin{tabular}{llccc}
\hline
Material & $E_0$ & $E_{45}$ & $E_{90}$ \\
\hline
\begin{tabular}{@{}l@{}}TPU-coated Nylon, \\ Quest Outfitters\end{tabular} & 215 MPa & - & 201 MPa \\[4mm]
\begin{tabular}{@{}l@{}}Silicone-coated Nylon, \\ Rockywoods SCA\end{tabular} & 96.5 MPa & 4.05 MPa & 120 MPa \\
\end{tabular}
\end{table}
\begin{figure}[h]
    \centering
    \includegraphics[width=0.4
    \textwidth]{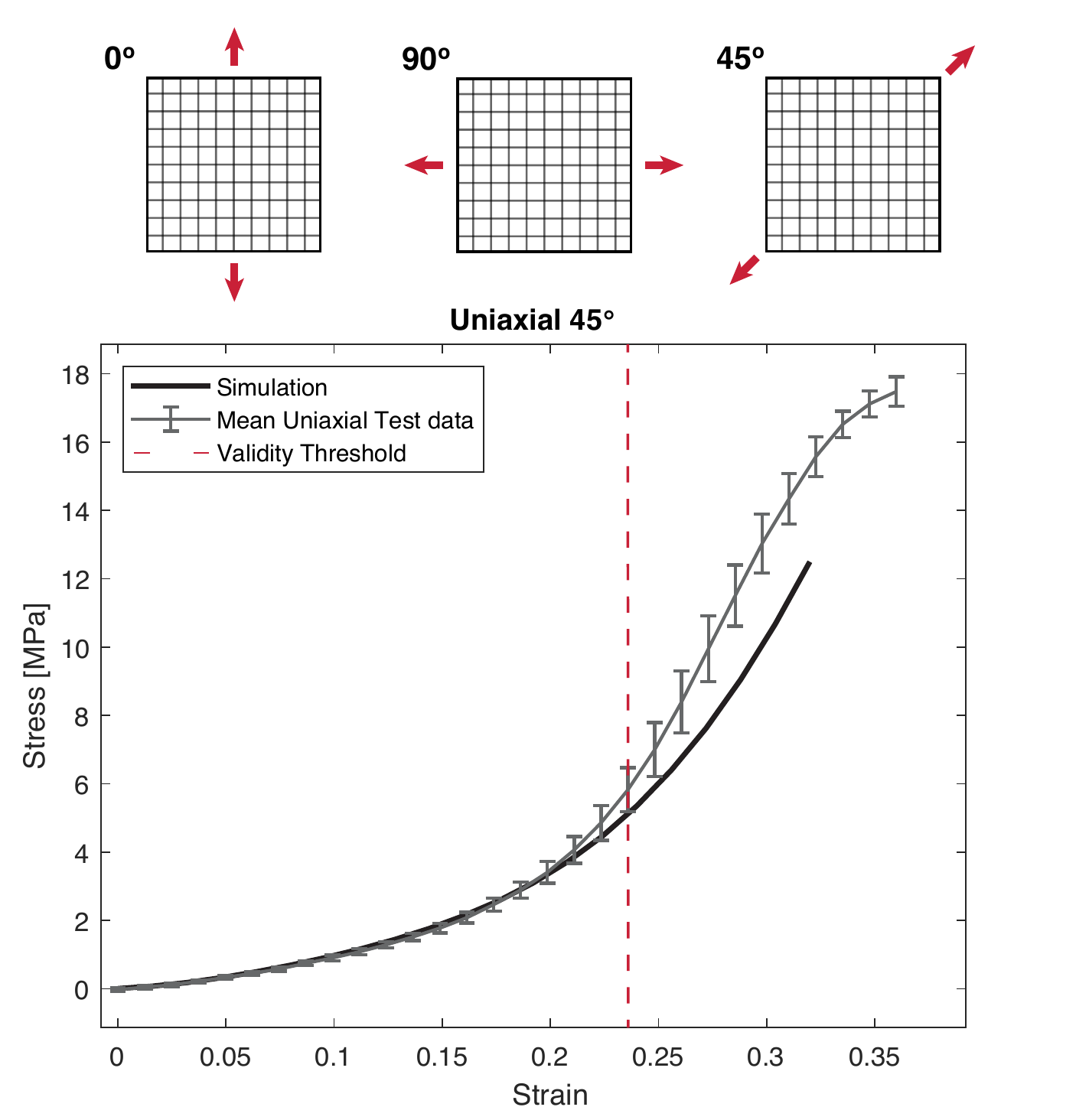}
    \caption{Top: uniaxial loading orientations; bottom: stress-strain curve for Silicone-coated TPU; the simulation data in black matches the test data until a validity threshold of roughly 24\% strain.}
    \label{fig:shear_fit}
\end{figure}
The main axis of the fPAM is cut at 45$^\circ$, because the actuation mechanism relies on anisotropy for the fPAM to contract and cause the IBR to bend.  
\newtext{Given that hyperelastic material models commonly used in FEA of soft robotics actuators (second order polynomial \cite{duPasquier2019DesignRobotics}, Yeoh \cite{Zhu22ModelAnalysis}, Ogden \cite{Moseley2016, Buffington20}) do not account for anisotropy \cite{Nguyen2020DesignDevices}, the material properties of the silicone-coated Nylon were modeled using the built-in \emph{*Fabric} material model in ABAQUS/Explicit}. \emph{*Fabric} is a data-based model that uses uniaxial test data along the fill and warp directions of a fabric, and its shear response. The experimental data acquired as described above was thus implemented directly into Abaqus for the fill and warp directions (0$^\circ$ and 90$^\circ$). The shear stress-strain curve was calculated by using the 45$^\circ$ uniaxial test data as a bias-extension test data (a common alternative to the picture-frame test) \cite{Launay2008ExperimentalReinforcements}. At the center of the test specimens, the material is in pure shear and the shear angle is directly related to \newtext{$d$}, the displacement during the uniaxial test:
\begin{equation}
    \gamma = \frac{\pi}{2}-2 \arccos\left(\frac{L_0+d}{\sqrt{2L_0}}\right),
\end{equation}
where $\gamma$ is the shear angle and $L_0$ is the initial specimen length. The shear force can be fit to uniaxial test data by relating it to the shear angle and $d$ using the following equation:
\begin{figure*}[h!]
    \centering
    \includegraphics[width=0.9  
    \textwidth]{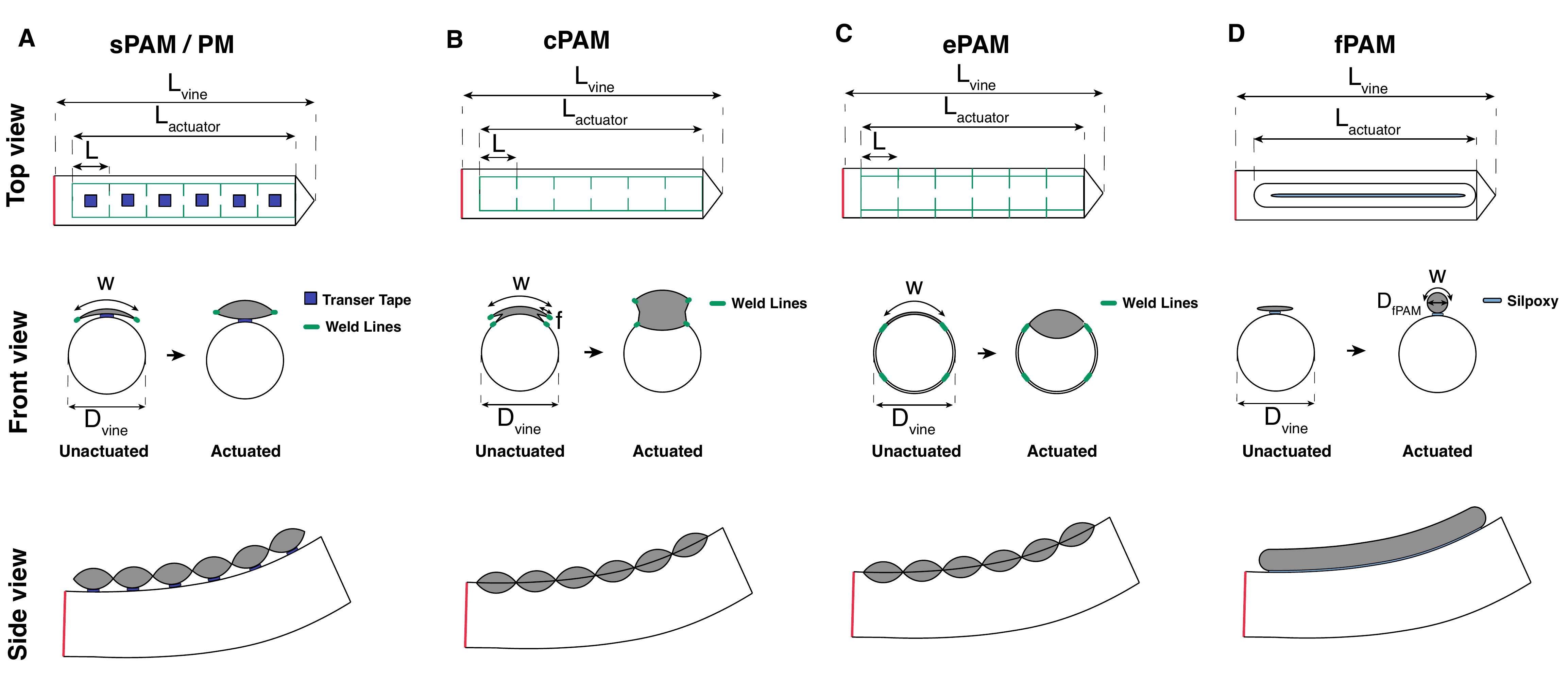}
    \caption{Inflated beam robot (IBR) fabrication and dimensions overview; top to bottom: schematic top, front, and side view (actuated) and main dimensions for A. series pneumatic artificial muscle/pouch motor (sPAM/PM), sides welded over two layers and taped onto the IBR, B. compression pneumatic artificial muscle (cPAM), sides welded over four layers and directly onto the IBR, C. embedded pneumatic artificial muscle (ePAM), two cylinders welded in a pattern to integrate the actuator in the IBR surface, D. fabric pneumatic artifical muscle (fPAM), glued using Silpoxy; the bold red lines indicate a \emph{Pinned} boundary condition in the model; \newtext{all portions described as `Transfer Tape' and `Weld Lines' are defined in the model with \emph{Tie} constraints; the areas denoted as `Silpoxy' are modeled as a \emph{Composite} surface.}}
    \label{fig:VineDim}
\end{figure*}
\begin{equation}
\begin{aligned}
    F_{sh}(\gamma) &= \frac{1}{(2H-3W)\cos{\gamma}} \left( \left(\frac{H}{W}-1\right) \cdot F \cdot \right. \\
    &\quad \left(\cos{\frac{\gamma}{2}}-\sin{\frac{\gamma}{2}}\right) -W \cdot F_{sh} \cdot \left(\frac{\gamma}{2}\right) \cdot \cos{\frac{\gamma}{2}} \left. \vphantom{\frac{H}{W}-1} \right),
\end{aligned}
\end{equation}
where \newtext{$H$ and $W$ are dimensions of the sample along and across loading-direction respectively, and $F$} is the external load applied during the uniaxial test. The fitting results are shown in Fig. \ref{fig:shear_fit}. The model matches the 45$^\circ$ test data well up to roughly 24\% strain; values beyond that differ due to the assumption of thread inextensibility. \newtext{Preliminary tests on the isolated fPAM muscle (not attached to the inflated beam) showed that the maximum contraction under the maximum pressure (30 kPa) were under the threshold, ascertaining the validity of the model in these conditions.} 

\subsection{Actuator Designs and Mechanisms}
\newtext{The actuator dimensions relate to bending force and actuator footprint. Considering $Pressure = \frac{Force}{Area}$, same area actuators have comparable bending forces. Second, IBRs often use eversion actuation, where larger foodprints mean higher eversion pressure (e.g., a wider actuator is harder to evert). Comparing actuators of same footprint equals comparing actuators of same bending efficiency.} \par  
The dimensions are given in Table \ref{Tab:ActDim}, and the design, bonding method, actuation method, and physical prototypes are shown in Fig. \ref{fig:VineDim}. The fabrication steps followed previous work (sPAMs/Pouch Motors \cite{Niiyama2015PouchDesign}, cPAMs \cite{Kubler2023ASteering}, ePAMs \cite{Abrar2021HighlyStructure}, fPAM \cite{Naclerio2020SimpleMuscle}). For the sPAMs/PMs and the fPAM, the actuators are fabricated separately from the IBR body and then attached using adhesive transfer tape (3M, Saint Paul, Minnesota, United States) and Silpoxy  adhesive (Reynolds Advanced Materials, Broadview, Illinois, United States) respectively. For the cPAMs and the ePAMs, the actuators are integrated on the surface of the IBR. The cPAMs are three layers of fabric selectively welded to create a form of origami pouch that unfolds under pressure. The ePAMs are created by selectively welding two parallel cylindrical IBR bodies into an actuation pattern. All four mechanisms are shown in the Front view of Fig. \ref{fig:VineDim}. \par 

\begin{table}[h]
\caption{Dimensions for the sPAM/PM, cPAM, ePAM, and fPAM designs.}
\label{Tab:ActDim}
\renewcommand{\arraystretch}{1.2}
\small
\begin{tabular}{lcccc}
\hline
 & sPAMs/PMs & cPAMs & ePAMs & fPAM \\
\hline
$\text{L}_\text{IBR}$ & 360 mm & 360 mm & 360 mm & 360 mm \\
$\text{D}_\text{IBR}$ & 80 mm & 80 mm & 80 mm & 80 mm \\
$\text{L}_\text{actuator}$ & 60 mm & 60 mm & 60 mm  & 300 mm \\
W & 60 mm & 60 mm & 60 mm & 60 mm \\
f & - & 24 mm & - & - \\
\end{tabular}
\end{table}

\subsection{Experimental Evaluation}
\begin{figure}[h]
    \centering
    \includegraphics[width=0.38
    \textwidth]{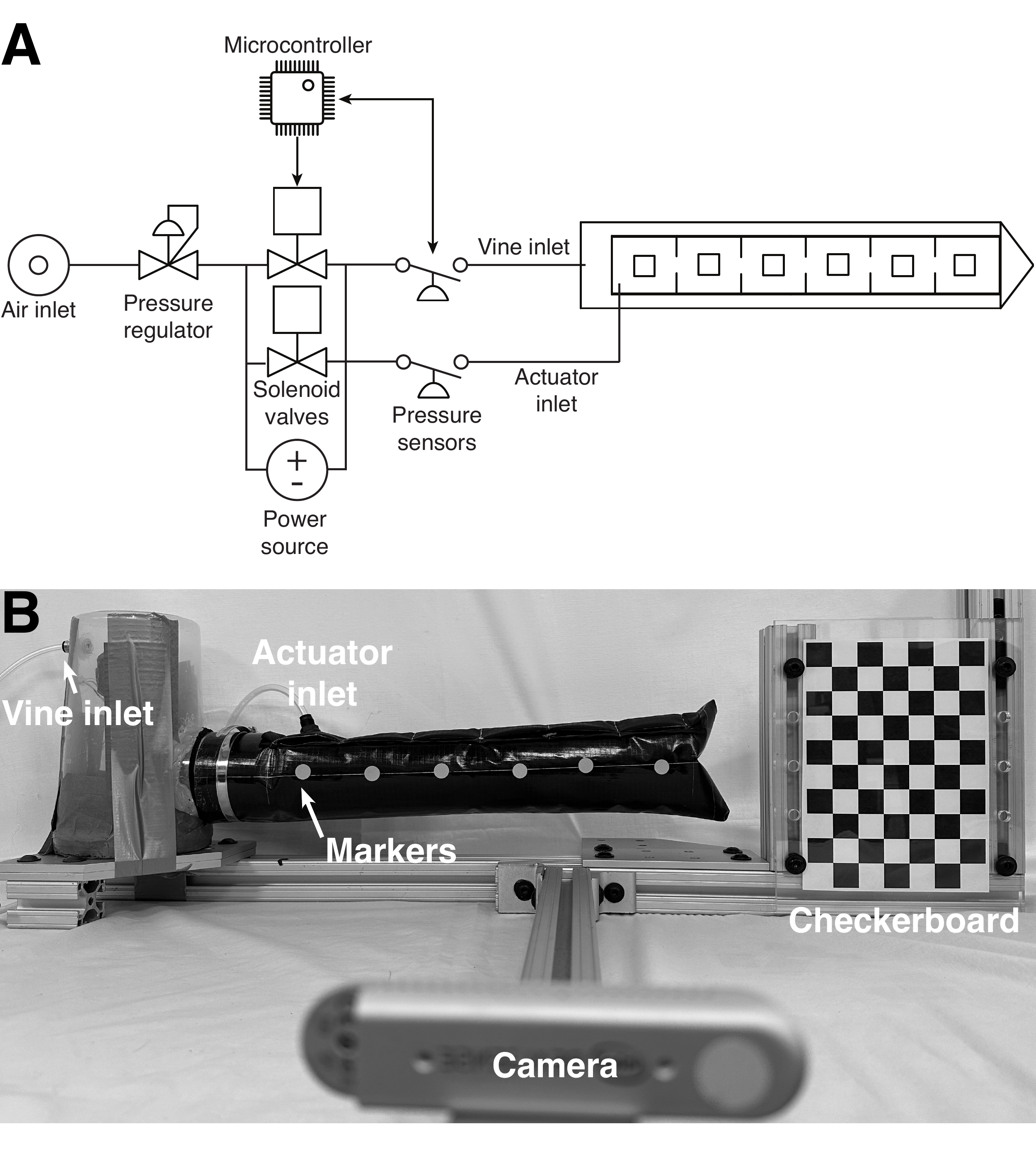}
    \caption{Experimental setup: A. schematic pressure control system; B. IBR displacement measurement setup, including both air inlets (left), camera (middle front), and checkerboard (right) for measurement calibration; white dots show the six markers used for displacement measurement.}
    \label{fig:setup}
\end{figure}

The experimental setup used to capture bending of the IBR and actuators is shown in Fig. \ref{fig:setup}. The control system, in Fig. \ref{fig:setup}A, is attached to a wall pressurized air inlet regulated manually to 100 kPa. The air is then split into two QB3 solenoid valves that regulate pressure through digital sensors placed close to the IBR and the actuators. A microcontroller (Arduino Uno) records the pressure measurements to the valves and controlled the valves based on a predetermined test sequence. An Intel Realsense D145 camera captures RGB and depth images, using a checkerboard for calibration. Six markers placed along the IBR and at the height of the middle of each actuator pouch are used to measure displacement. The data acquired by the camera is post-processed using Python and plotted using Matlab. \par
Previous work showed that gravity has only a negligible effect on the measurements due to the light weight of the materials used in these experiments \newtext{\cite{Kubler2023ARobots}}. The experimental and simulation setup are otherwise built to be as physically similar as possible. \par 

Each actuator type was tested using the protocol below:
\begin{enumerate}
    \item IBR pressurized at 2 kPa
    \item Actuator sequentially pressurized to 10, 20, and 30 kPa consecutively for 5 seconds each
    \item Actuator depressurized for 5 seconds
\end{enumerate}
The 5-second duration for the inflation and deflation allow for the pressure in the actuators to achieve steady-state, which provides a more accurate reading of the pressure/displacement relationship. It is also close to what the FE procedure simulates, where each pressure is simulated and held to a threshold to avoid inertial effects.

\subsection{Metrics}
The finite element model and experimental displacement data are compared based on the XY-displacement of the markers. First, using a \emph{Path} along the center line of the body of the IBR, the XY displacement coordinates vs. pressure are extracted from the output database (ODB) of the simulations. They are then plotted against the measured marker displacements at P = 10, 20, 30 kPa, using the pressures measured at the inlet for the actuators and the final simulation pressure. The accuracy is calculated as one minus the least square error between the marker positions ($y_\text{EXP}$) and the simulation displacement curve ($y_\text{FEA}$): \\ 
\begin{equation}
    a = 1 - \frac{1}{N_\text{markers}} \sum_{n=1}^{N_\text{markers}} \left( \frac{y_\text{FEA} - y_\text{EXP}}{y_\text{FEA}} \right)^2 
\end{equation} 

\section{Results and Discussion}
\begin{figure*}[h!]
    \centering
    \includegraphics[width=0.9
    \textwidth]{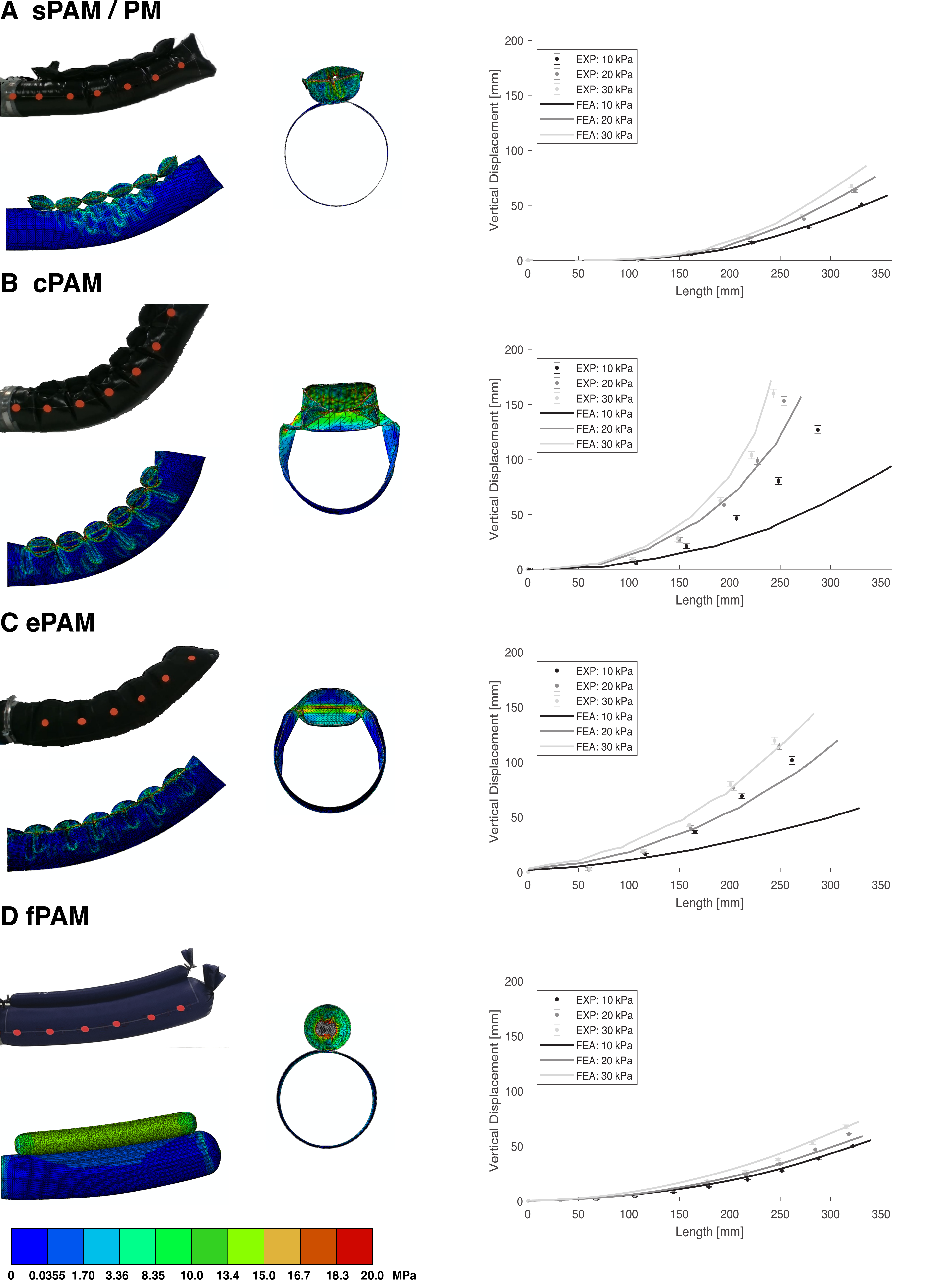}
    \caption{Results and comparison of experimental and FEA results for the four actuator types; in each actuator section there is a photograph of the actuator at 30 kPa (top, left), a capture of the von Mises Stress in FEA at 30 kPa from the side and a cross-section (center), and a graph comparing the displacements of the experiments (mean) and FEA in length and height at 10, 20, and 30 kPa (right); the experimental error bar represents the variation from the mean; the stress scale at the bottom is the same for all FEA results and capped at 20 MPa. }
    \label{fig:Res}
\end{figure*}
\begin{table}
\caption{Accuracy of FE model and maximum experimental vertical displacement for the \\ sPAM/PM, cPAM, ePAM, and fPAM actuators}
\center
\renewcommand{\arraystretch}{1.5}
\label{Tab:Error}
\scalebox{1.2}{
\begin{tabular}{lcccc}
\hline
 & 10 kPa & 20 kPa & 30 kPa & $\text{d}_\text{max}$ \\
\hline
sPAMs/PMs & 96.4\% & 95.8\% & 94.3\% & 67.5 mm\\
cPAMs & 87.1\% & 97.0\% & 97.3\% & 159.7 mm\\
ePAMs & 80.7\% & 92.1\% & 94.9\% & 119.4 mm  \\
fPAM & 97.6\% & 97.3\% & 96.7\% & 67.2 mm\\
\end{tabular}}
\end{table}
In this section, we first compare experimental and FEA results for individual actuators, then compare the actuators to each other. We discuss differences bending mechanisms, bending range, and their effect on the accuracy of the simulation. We discuss the FEA strategy and its implications for the results. Finally, we analyze how the FEA presented in this work contributes to the fields of IBRs and Soft Robotics in terms of design and control. 
\newtext{\subsection{sPAMs}}
The results for each actuator are shown in Fig. \ref{fig:Res}, and FEA accuracy values and maximum displacements (at 30 kPa) are given in Table \ref{Tab:Error}. Starting with the sPAMs (Fig. \ref{fig:Res}.A), the experimental and simulation data correspond very well, with a minimum accuracy of 94.3\%. The actuators are external to the IBR body so their axis of contraction is further away from the IBR central axis, which in turn limits the bending radius. The actuator location also implies that the IBR cross-section is affected very little by bending. The stress peaks, very similar in location and magnitude to those of the ePAMs, are at the weld seams. The buckling pattern is irregular compared to the cPAMs and ePAMs, which indicates that the effect of external actuators is less evenly distributed and limits the bending range. \par
\begin{figure}[t]
    \centering
    \includegraphics[width=0.4
    \textwidth]{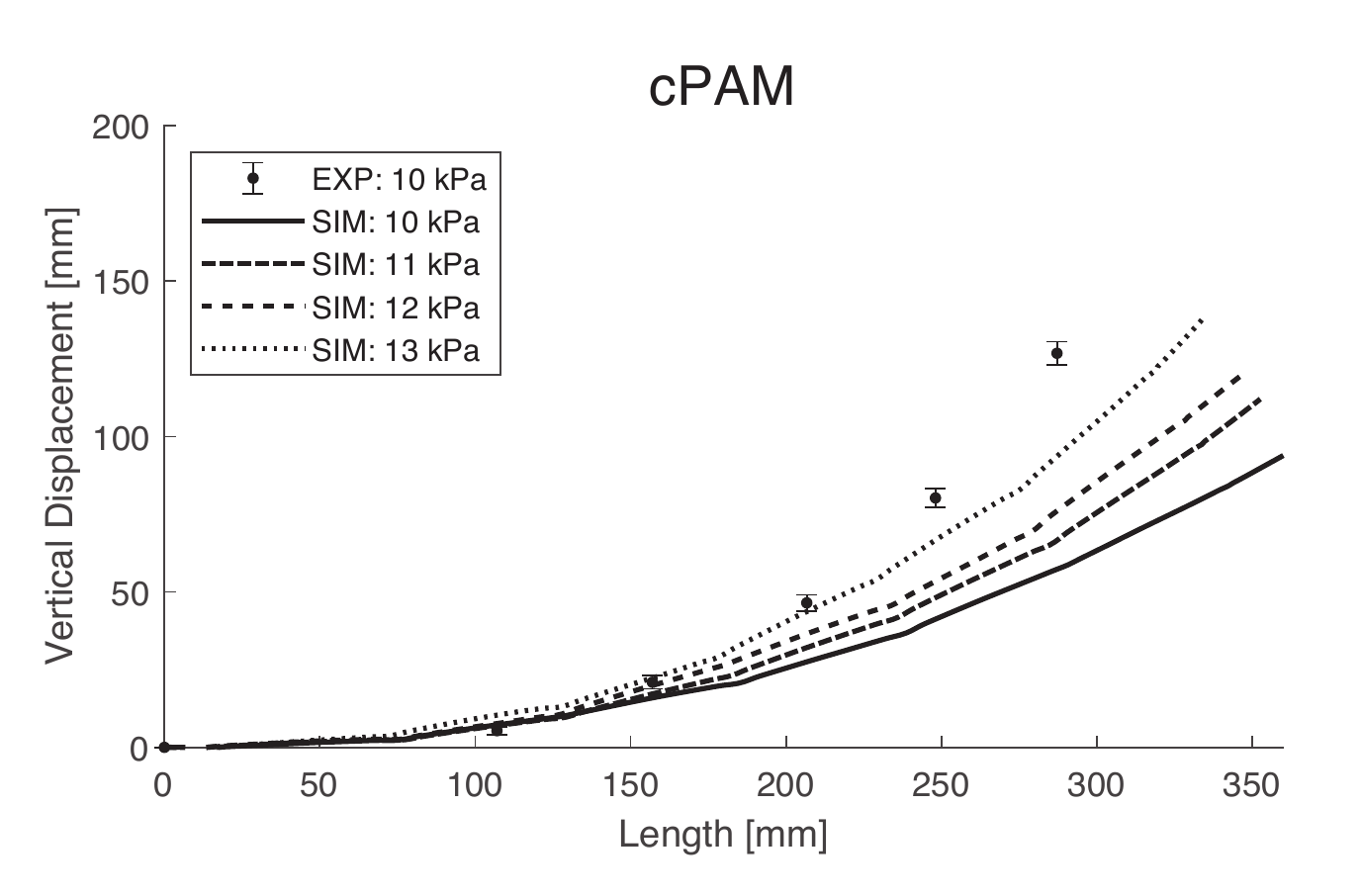}
    \caption{FEA simulation values for the cPAM at 11, 12, and 13 kPa to test the hypothesis that the FEA has lower accuracy at partially buckled states.}
    \label{fig:cPAM}
\end{figure}

\newtext{\subsection{cPAMs}}
The cPAMs (Fig. \ref{fig:Res}.B) display the most substantial displacement and pronounced curvature. Their maximum vertical displacement is 1.34 times that of ePAMs and 2.37 times that of sPAMs/fPAM. Embedded at the surface like the ePAMs, the cPAMs are closer to the center axis of the IBR than the sPAMs and fPAM.  The added fabric folds lead to a 2.60-fold increase in their inner area compared to ePAMs. These folds contribute to a dual buckling pattern: one between the actuators (similar to ePAMs) and another at the actuator itself. Stress peaks primarily emerge at weld lines, especially where four fabric layers connect. The FEA accuracy of the cPAMs exceeds 97\% for pressures over 20 kPa but drops by 10\% for lower pressures due to buckling-induced stress singularities. This inherently unstable mechanism induces sudden and substantial bending changes. At 10kPa, when the buckling pattern is only partially formed, deformation is unstable and very sensitive to slight pressure variations. To support this hypothesis, we show the cPAMs deformation at 11, 12, and 13 kPa in Fig. \ref{fig:cPAM}. A 3 kPa pressure difference increases maximum vertical displacement by nearly 30\%. Given that the experimental setup includes a margin of error for pressure control, this explains the lower accuracy at 10 kPa. 
\newtext{\subsection{ePAMs}}
The ePAMs (Fig. \ref{fig:Res}.C) combine bending elements from the sPAMs and cPAMs. The deformed actuator shape is the same as the sPAMs/PMs, and, as for the cPAMs, the actuators are embedded directly on the IBR and rely on a regular buckling pattern to deform. The displacement range of the ePAMs thus lies between the two other actuator types. Because the bending mechanism is the same as the cPAMs, the same stress singularities are found at low pressures ($<$20 kPa) and partially buckled states. Although the ePAMs are slightly less accurate than the cPAMs in our experiments, the fabrication of ePAMs is more straightforward because it only involves welding two surfaces (no extra folds), and they can easily be combined on the IBR surface for 3D control. 
\newtext{\subsection{fPAMs}}
Finally, the fPAM (Fig. \ref{fig:Res}.D) has the smallest bending range of the four, given that the bending mechanism relies solely on material contraction and that the actuator axis is furthest away from the IBR center axis. However, the bending mechanism is the most stable, and the FEA corresponds to the experiments with over 96.7\% accuracy for all pressures. The fabric model in Abaqus causes oscillations in the results that can be mitigated by controlling the simulation time step. The deformation amplitude is not affected, but the computational cost of the FEA increases with increasing time step. A judicious choice of time step and mass scaling will affect the run time and the oscillation frequency. We find that the oscillations always have the same amplitude, which corresponds to the experimental deformed state. 
\newtext{\subsection{General Considerations}}
The mesh size significantly affects the accuracy of Finite Element Analysis (FEA), especially for cPAMs and ePAMs that rely on buckling for bending. We tested mesh sizes from 0.5 to 4.0 mm. While ePAMs and fPAMs were not sensitive, accuracy for cPAMs and ePAMs increased by almost 20\%. Finer mesh led to deeper buckling patterns, increasing bending radius. However, finer mesh also exponentially increased run times, so our chosen mesh balances accuracy and efficiency. Users can adjust mesh size based on needing qualitative insights or quantitative data for designing purposes.\par
Overall, the FEA results confirm the accuracy and potential of using FEA in IBR design and deployment. Actuators differentiate based on bending mechanism (geometry-based or material-based contraction), external/embedded system, and inner surface area. Combining geometry and embedded actuation with a large surface area provides the broadest deformation range but also shows more buckling and inaccuracies at lower pressures. Material-based external actuation is highly accurate but with a limited range. These insights help roboticists pick optimal designs. For intricate path navigation without much environment interaction, cPAMs excel. For minor obstacle navigation and tip steering, fPAMs are simpler. \par
FEA is valuable for aiding IBR fabrication in multiple aspects. For instance, the von Mises stress distribution can highlight potential rupture zones in a design. If weld strength is known, ruptures can be predicted and managed via design tweaks, material adjustments, or refining weld parameters computationally. This accelerates the process, saving time and resources. Our model stands out by offering this kind of support, motivating us to share it openly online for others to use and improve, even introducing new actuator types.\par
An additional advantage of using FEA modeling in IBR design is that very long IBR models can be easily built and tested virtually. While other models tend to mimic isolated IBR deformation, FEA allows interactions with the environment through additional steps and timed external loads. By simulating an environment virtually, the actuator choice and design can be tailored for specific applications or experimental challenges an IBR might face. \newline

\section{Conclusion}
This work validates a new approach of modeling pneumatically-actuated inflated beam soft robots, or IBRs using FEA. The \emph{Dynamic, Explicit} formulation of FEA accurately captures complex buckling patterns and anisotropic material models, suitable both geometry-based and material-based bending mechanisms. The model converges accurately for the four primary types of pneumatic actuators used with IBRs. \par
Combining buckling-based and embedded actuator design yields in the highest bending curve and deformation range. However, stress singularities during buckling at lower pressures affect model accuracy. Material-based deformation retains over 96.7\% accuracy at all pressures but has limited deformation potential. \par 
The FEA approach proposed here can be adapted to achieve qualitative and quantitative results that streamline different stages of the IBR design process. This tool is crucial for enhancing IBR efficiency, avoiding wasteful ruptures or failures, and by tailoring actuator use and positioning to specific applications.

\bibliographystyle{IEEEtran}
\bibliography{references_nm}

\ifCLASSOPTIONcompsoc
  \section*{Acknowledgments}
\else
  \section*{Acknowledgment}
\fi

The authors thank Alexander Kübler for the experimental procedure and measurements performed with the sPAM/PM, cPAM, and fPAM.

\ifCLASSOPTIONcaptionsoff
  \newpage
\fi




%

\end{document}